\documentclass[conference]{IEEEtran}
\IEEEoverridecommandlockouts

\usepackage{cite}
\usepackage{amsmath,amssymb,amsfonts}
\usepackage{algorithmic}
\usepackage{graphicx}
\usepackage{textcomp}
\def\BibTeX{{\rm B\kern-.05em{\sc i\kern-.025em b}\kern-.08em
    T\kern-.1667em\lower.7ex\hbox{E}\kern-.125emX}}
\usepackage{times}
\usepackage{latexsym}
\usepackage{graphicx}
\usepackage{color}
\usepackage{multirow}
\usepackage{amsmath}
\usepackage{amssymb}
\usepackage[usenames,dvipsnames]{xcolor}
\usepackage[ruled,vlined,linesnumbered]{algorithm2e}
\usepackage{float}
\usepackage{tabularx}
\usepackage{multicol}

\newtheorem{definition}{Definition}

\newcommand{\comment}[1]{}

\makeindex
\begin{document}

\title{Finding Optimal Solutions to Token Swapping by Conflict-based Search and Reduction to SAT
}

\author{\IEEEauthorblockN{Pavel Surynek}
\IEEEauthorblockA{\textit{Faculty of Information Technology} \\
\textit{Czech Technical University in Prague}\\
Czech Republic \\
pavel.surynek@fit.cvut.cz}
}

\maketitle
\begin{abstract}
We study practical approaches to solving the {\em token~swapping} (TSWAP) problem optimally in this short paper. In TSWAP, we are given an undirected graph with colored vertices. A colored token is placed in each vertex. A pair of tokens can be swapped between adjacent vertices. The goal is to perform a sequence of swaps so that token and vertex colors agree across the graph. The minimum number of swaps is required in the optimization variant of the problem.
We observed similarities between the TSWAP problem and {\em multi-agent path finding} (MAPF) where instead of tokens we have multiple agents that need to be moved from their current vertices to given unique target vertices. The difference between both problems consists in local conditions that state transitions (swaps/moves) must satisfy.
We developed two algorithms for solving TSWAP optimally by adapting two different approaches to MAPF - CBS and MDD-SAT. This constitutes the first attempt to design optimal solving algorithms for TSWAP. Experimental evaluation on various types of graphs shows that the reduction to SAT scales better than CBS in optimal TSWAP solving.
\end{abstract}


\section{Introduction}
\noindent \noindent


The {\em token swapping problem (TSWAP)} (also known as {\em sorting on graphs}) \cite{DBLP:conf/fun/YamanakaDIKKOSSUU14} represents a generalization of sorting problems a fundamental task in computer science \cite{DBLP:journals/jal/Thorup02}. While in the classical sorting problem we need to obtain linearly ordered sequence of elements by swapping any pair of elements, in the TSWAP problem we are allowed to swap elements at selected pairs of positions only. Usually the minimum number of swaps is desirable both in the classical sorting and in the TSWAP problem.

\begin{figure}[t]
    \centering
    \includegraphics[trim={5.2cm 23cm 5.2cm 3.7cm},clip,width=0.55\textwidth]{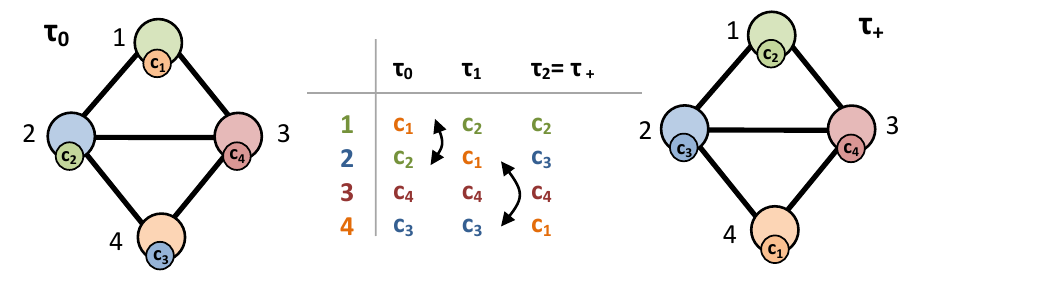}
    \vspace{-0.8cm}\caption{A TSWAP instance. A solution consisting of two swaps is shown.}
    \label{figure-TSWAP}
\end{figure}

Using a modified notation from \cite{DBLP:journals/tcs/YamanakaDIKKOSS15} the TSWAP problem is given by an undirected graph $G=(V,E)$ with vertex set $V$ and edge set $E$. Each vertex in $G$ is assigned a color in $C = \{c_1,c_2,...,c_h\}$ via $\tau_+:V\rightarrow C$. A token of a color in $C$ is placed in each vertex. The task is to transform a current token placement into the one such that colors of tokens and respective vertices of their placement agree. Desirable token placement can be obtained  by swapping tokens on adjacent vertices in $G$. See Figure \ref{figure-TSWAP} for an example instance of TSWAP.

\begin{figure}[h]
    \centering
    \includegraphics[trim={5.2cm 23cm 4.8cm 3.5cm},clip,width=0.55\textwidth]{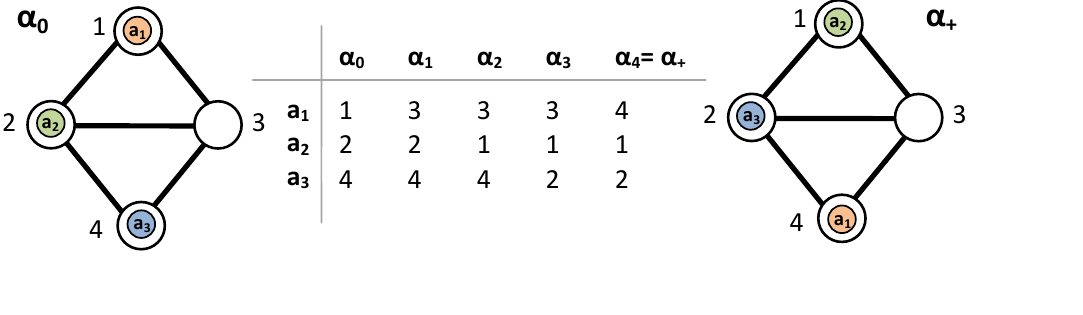}
    \vspace{-1.0cm}\caption{A MAPF instance with three agents $a_1$, $a_2$, and $a_3$.}
    \label{figure-MAPF}
\end{figure}

A {\em multi-agent path finding} (MAPF) problem \cite{DBLP:conf/aiide/Silver05,DBLP:journals/jair/Ryan08} is similar to TSWAP. MAPF consists of an undirected graph $G=(V,E)$ and a set of agents $A=\{a_1, a_2, ..., a_k\}$ such that $|A|<|V|$. Each agent is placed in a vertex so that at most one agent resides in each vertex. The placement of agents is denoted $\alpha: A \rightarrow V$.  The task is to move agents in a non-conflicting way so that each agent reaches its goal vertex. An agent can move into adjacent unoccupied vertex provided no other agent enters the same target vertex. An example of MAPF instance is shown in Figure \ref{figure-MAPF}.

The TSWAP problem has been introduced recently. So far theoretical results concerning computational complexity \cite{DBLP:conf/stacs/BonnetMR17} and approximations \cite{DBLP:conf/esa/MiltzowNORTU16,DBLP:conf/walcom/YamanakaDHKNOSS17} have appeared. To our best knowledge there is no study dealing with practical solving of TSWAP optimally. We would like to start to fill in this gap in this short paper.

The similar MAPF problem has been studied longer and many theoretical results \cite{DBLP:journals/jsc/RatnerW90} as well as practical solving algorithms exist for MAPF \cite{standley2010finding,DBLP:conf/ictai/Surynek14,SharonSFS15,DBLP:journals/ai/SharonSGF13}. Hence we would like advance knowledge in solving both problems by initiating their cross-fertilization. Our current contribution consists in adapting optimal MAPF solvers for the TSWAP problem. We adapted Conflict-Based Search (CBS) \cite{SharonSFS15} and MDD-SAT \cite{SurynekFSB16} based on reduction to {\em propositional satisfiability} \cite{Biere:2009:HSV:1550723}.

\section{Background}

We denote by $\tau:V\rightarrow C$ colors of tokens placed in vertices of $G$. That is, $\tau(v)$ for $v \in V$ is a color of a token placed in $v$. Starting placement of tokens is denoted as $\tau_0$; the goal token placement corresponds to $\tau_+$. Transformation of one placement to another is captured by the concept of {\em adjacency} defined as follows \cite{DBLP:journals/tcs/YamanakaDIKKOSS15,DBLP:conf/walcom/YamanakaDHKNOSS17}:

\begin{definition}
    Token placements $\tau$ and $\tau'$ are said to be adjacent if there exist a subset of non-adjacent edges $F \subseteq E$  such that $\tau(v) =  \tau'(u)$ and $\tau(u) = \tau'(v)$ for each $\{u,v\} \in F$ and for all other vertices $w \in V \setminus \bigcup_{\{u,v\} \in F}{\{u,v\}}$ it holds that $\tau(w) = \tau'(w)$. \footnote{The presented version of adjacency is sometimes called {\em parallel} while a term adjacency is reserved for the case with $|F|=1$.}
    \label{def:adjacency}
\end{definition}

The task in TSWAP is to find a swapping sequence of token placements $[\tau_0, \tau_1, ..., \tau_m]$ such that $\tau_m = \tau_+$ and $\tau_i$ and $\tau_{i+1}$ are adjacent for all $i=0,1,..., m-1$. It has been shown that for any initial and goal placement of tokens $\tau_0$ and $\tau_+$ respectively there is a swapping sequence transforming $\tau_0$ and $\tau_+$ containing $\mathcal{O}({|V|}^2)$ swaps \cite{YamanakaComplex2016}. The proof is based on swapping tokens on a spanning tree of $G$. Let us note that the above bound is tight as there are instances consuming $\Omega({|V|}^2$) swaps. It is also known that finding swapping sequence that has as few swaps as possible is an NP-hard problem.

Similarly in MAPF we are given initial configuration of agents $\alpha_0$ and goal configuration $\alpha_+$. At each time step an agent can either {\em move} to an adjacent location or {\em wait} in its current location. The task is to find a sequence of move/wait actions for each agent $a_i$, moving it from $\alpha_0(a_i)$ to $\alpha_+(a_i)$ such that agents do not {\em conflict},
i.e., do not occupy the same location at the same time.

\begin{definition}
    Configuration $\alpha'$ results from $\alpha$ if and only if the following conditions hold: (i) $\alpha(a) = \alpha'(a)$ or $\{\alpha(a),\alpha'(a)\} \in E$ for all $a \in A$ (agents wait or move along edges); (ii) for all $a \in A$ it holds that if ${\alpha(a) \neq \alpha'(a)} \Rightarrow {\alpha'(a) \neq \alpha(a')}$ for all $a' \in A$ (target vertex must be empty); and (iii) for all $a,a' \in A$ it holds that if ${a \neq a'} \Rightarrow {\alpha'(a) \neq \alpha'(a')}$ (no two agents enter the same target vertex).
    \label{def:movement}
\end{definition}

Solving the MAPF instance is to search for a sequence of configurations $[\alpha_0,\alpha_1,...,\alpha_{\mu}]$ such that  $\alpha_{i+1}$ results using valid movements from $\alpha_{i}$ for $i=1,2,...,\mu-1$, and $\alpha_{\mu}=\alpha_+$. 

MAPF is usually solved aiming to minimize one of the two commonly-used global
cumulative cost functions: (1) Sum-of-costs is the summation, over all agents, of the number of time steps required to reach the goal location~\cite{DBLP:journals/ai/SharonSGF13}. (2) Makespan: is the time until the last agent reaches its destination (i.e., the maximum of the individual costs)~\cite{DBLP:conf/ictai/Surynek14}.

\section{Related Work}

In contrast to the upper bound complexity result for the TSWAP problem, it can be shown that any solvable MAPF instance can be solved using $\mathcal{O}({|V|}^3)$ moves \cite{DBLP:conf/focs/KornhauserMS84,luna2011efficient}. Moreover this a tight bound again as there as instances that need $\Omega({|V|}^3)$ moves.
As MAPF is of great practical interest many optimal, sub-optimal, and bounded sub-optimal solvers have been developed for MAPF. Our scope here is limited to optimal solvers. Optimal solvers for MAPF can be divided to two classes. (1) Search-based solvers. These algorithms consider MAPF as a graph search problem. Some of these algorithms are variants of the A* algorithm that search in a global {\em search space} -- all different ways to place $k$ agents into $V$ vertices, one agent per vertex~\cite{standley2010finding,DBLP:journals/ai/WagnerC15}.
Others algorithms such as ICTS~\cite{DBLP:journals/ai/SharonSGF13} and CBS~\cite{SharonSFS15} search different search spaces and employ novel (non-A*) search tree. (2) Reduction-based solvers. By contrast, many recent optimal solvers reduce MAPF to known problems such as CSP~\cite{DBLP:conf/icra/Ryan10}, SAT~\cite{surynek2012towards}, Inductive Logic Programming~\cite{yu2013planning} and Answer Set Programming~\cite{erdem2013general}.

\section{Adapting Algorithms for TSWAP}

We modified two optimal MAPF algorithms: CBS \cite{SharonSFS15} and MDD-SAT \cite{SurynekFSB16} for the TSWAP problem. Both algorithms require modifications in handling of state transitions.

\subsection{Adapting the CBS Algorithm}

The first algorithm undergoing our modifications is CBS. The algorithm is based on resolving collisions between agents via adding constraints that forbid occurrence of resolved collisions. The top level search of CBS uses priority queue that stores partial solutions together with a set of {\em conflicts}. The priority is determined by the value of objective for a partial solution. In addition to this, CBS has a {\em low level search} that basically finds shortest paths connecting agent's initial positions and goals while ignoring collisions between agents. The low level search uses a set of conflicts that the path has to avoid. Conflicts are triples $(a, v, t)$ with $a \in A$, $v \in V$, and timestep $t$ which means that the path being searched for agent $a$ must avoid $v$ at timestep $t$.

Initially, we use shortest paths without considering any conflicts as an initial partial solution and store it into the priority queue. The top level search always takes the partial solution $P$ with the lowest value of the objective and set of conflicts $P.conflicts$ associated with $P$. $P$ is then checked for collisions between agents. If there are no collisions, then we can return $P$ as an optimal solution. Otherwise we take the first collision; let this collision happened between agents $a_i$ and $a_j$ in vertex $v$ at timestep $t$. Collision $(a_i, a_j, v, t)$ will be resolved via branching at the top level search. One or the other involved agent must avoid $v$ at $t$. This is carried out by adding two new states into the priority queue: partial solutions obtained by the low level search with respect to $P.conflicts \cup \{(a_i, v, t)\}$ and $P.conflicts \cup \{(a_j, v, t)\}$ associated with their extended conflict sets. For details on CBS we refer the reader to \cite{SharonSFS15}.

\subsubsection{Considering TSWAP Specific Collisions}

In TSWAP collisions between tokens are understood differently than in MAPF. A collision between tokens $t_i$ and $t_j$ occurs in the following cases:

\begin{itemize}
\item{tokens $t_i$ and $t_j$ both attempt to occupy vertex $v$ at time step $t$}
\item{token $t_i$ enters a vertex $v$ from vertex $u$ at timestep $t$ but $t_j$ appearing in $u$ at timestep $t$ is different from a token residing previously in $v$ at timestep $t-1$}
\end{itemize}

Both cases represent a collision $(t_i, t_j, v, t)$ that is resolved by introducing conflicts $(t_i, v, t)$ and $(t_j, v, t)$ and by updating partial solutions accordingly.

\subsection{Adapting the MDD-SAT Algorithm}
The second algorithm we adapted is a compilation-based MDD-SAT \cite{SurynekFSB16}. This algorithm reduces an instance of MAPF to a series of decision Boolean satisfiability (SAT) \cite{Biere:2009:HSV:1550723} instances. Versions of MDD-SAT optimizing various objectives such as the {\em makespan} or the {\em sum-of-costs} in MAPF exist. The crucial step of the approach builds on formulating the decision of whether there is a solution to given MAPF of a specified value of the objective as an instance of SAT. Our major task when adapting MDD-SAT for TSWAP is hence to modify the SAT formulation to reflect different state transitions of TSWAP.

The SAT model of MAPF in MDD-SAT relies on a {\em time expansion} of $G$. That is we have a copy of $G$ for every timestep \cite{DBLP:journals/amai/Surynek17}. Search for plans for individual agents can then be interpreted as a search for non-conflicting paths in the time expanded graph (an agent can visit a vertex multiple times hence plans for individual agents cannot be easily expressed as simple paths in unexpanded graph).

Boolean variables $\mathcal{X}(a)_v^t$ for each $v \in V$, $a \in A$ and each timestep $t$ modeling occurrence of agent $a$ in vertex $v$ at timestep $t$ are introduced. Auxiliary Boolean variables $\mathcal{E}(a)_{u,v}^t$ representing traversal of edge $\{u,v\}$ by agent $a$ at timestep $t$ are typically used for easier expressing of constraints. Constraints are introduced to restrict possible assignments of values to $\mathcal{X}(a)_v^t$ and $\mathcal{E}(a)_{u,v}^t$ variables so that assignments represent only valid solutions to MAPF - see for details \cite{SurynekFSB16}.

We will follow similar scheme in the TSWAP problem. We introduce Boolean variable $\mathcal{Y}(c)_v^t$ representing occurrence of a token of color $c$ in vertex $v$ at time step $t$ and analogically $\mathcal{S}(c,d)_{u,v}^t$ for $c,d \in C$ representing swap at edge $\{u,v\}$ involving colors $c$ and $d$ (color $c$ starts in $u$). Most of constraints such as those enforcing that only one color can assigned to each vertex can be taken directly from the MAPF encoding, since analogical properties must hold for agents. However there are some differences in TSWAP that need to be treated specifically.

\subsubsection{Considering TSWAP Specific Constraints}

For instance, swapping tokens along edge $\{u,v\}$ at time step $t$ is slightly different as in MAPF we move an agent into a vacant vertex while in TSWAP we need to replace an outgoing token with that from the target vertex:

\[ \mathcal{S}(c,d)_{u,v}^t \Rightarrow \mathcal{Y}(c)_v^t \wedge \neg \mathcal{Y}(c)_v^{t+1} \wedge \neg \mathcal{Y}(d)_v^{t} \wedge \mathcal{Y}(d)_v^{t+1}~~~~(1)\]

At any time step $t$ assignment of variables must encode a valid placement of tokens in vertices of the graph. So at most one color is assigned to each vertex ensured by the following constraint for each vertex $v$ and timestep $t$.

\[ \sum_{\{c\} \in C} {\mathcal{Y}(c)_{v}^t} \leq 1 \]

To illustrate other constraints we show how to enforce non-conflicting swaps of tokens. This can be expressed by {\em at-most-one } constraint over edge variables as follows for a fixed vertex $u$ and colors $c,d$. This constraint ensures that at most one swap occurs in a vertex a given timestep.

\[ \sum_{\{u,v\} \in E} {\mathcal{S}(c,d)_{u,v}^t} \leq 1 \]

The suggested encoding permits multiple non-conflicting swaps per single timestep (that is, $|F|>1$ in the Definition \ref{def:adjacency}). Observe that rotations of tokens over {\em non-trivial cycles} (a swap is a trivial rotation over an edge) consisting of at least three edges, that is common in some variants of MAPF, is forbidden by the encoding as it would violate constraint (1). Chain like movements from MAPF, where only the leader of a chain of agents need to enter vacant vertex while all other agents of the chain enter a vertex being simultaneously vacated by the preceding agent, are also forbidden.

\subsubsection{Considering TSWAP Specific Objective}

In TSWAP, we minimize the number of swaps which is an objective different from both the makespan and the sum-of-costs being implemented in MDD-SAT. As the SAT-based approach always answers a given formula in a yes/no manner we need to translate the minimization of the number of swaps in TSWAP to a series of queries to the SAT-solver. We always build a formula using above constructs that encodes a question whether there is a solution to TSWAP using specified number of swaps $\sigma$. By posting queries for $\sigma=1,2,3 ...$ one can obtain the minimum number of swaps as $\sigma$ corresponding to the first satisfiable formula.

In our practical implementation we do not start with $\sigma=1$ but with a lower bound estimation corresponding to the sum of lengths of shortest paths connecting the starting and target token positions divided by 2 (a single swap can move 2 tokens towards their target vertices each by one step forward).

Encoding the $\sigma$ bound can be done through various cardinality constraints in the formula \cite{DBLP:conf/cp/SilvaL07} on top of edge variables $\mathcal{S}(c,d)_{u,v}^t$. Once an edge variable is set to $True$ a corresponding swap is to be made. The following constraint hence need to be encoded in the formula through cardinality constraints.

\[ \sum_{\{u,v\} \in E, c,d \in C, c < d, t=1,2,...,\sigma} {\mathcal{S}(c,d)_{u,v}^t} \leq \sigma \]

As multiple token swaps can occur within the suggested encoding per single timestep, it is not necessary to make expansions for all steps up to $\sigma$. However careful setting of the number of time expansions with respect to given $\sigma$ need to be done. We need to use sufficient number of expansions to ensure that if there is a swapping sequence of length $\sigma$ we can find it within the given number of expansions. Precise calculation of the number of expansions depending on the cost parameter in MAPF has been done in \cite{SurynekFSB16}. We omit the analogical calculation for TSWAP here for the sake of brevity.

\subsubsection{Reducing the Number of Variables}

A common approach to reduce the number of decision variables in solving approaches to MAPF is the use of {\em multi-value decision diagrams} (MDDs) \cite{DBLP:journals/ai/SharonSGF13}. The basic observation that holds both for MAPF and TSWAP is that a token/agent can reach vertices in the distance $d$ (distance of a vertex is measured as the length of the shortest path) from the current position of the agent/token no earlier than in the $d$-th time step. Analogical observation can be made with respect to the distance from the goal position.

Above observations can be utilized when making the time expansion of $G$. For a given token, we do not need to consider all vertices at time step $t$ but only those that are reachable by the token in $t$ timesteps from its initial position and that ensure that token's goal can be reached in the remaining $\sigma - t$ timesteps. This idea can reduce the size the expanded graph significantly and consequently can reduce the size of the Boolean formula by eliminating $\mathcal{Y}(c)_v^t$ and $\mathcal{S}(c,d)_{u,v}^t$ variables correspoding to unreachable vertices.

\begin{figure}[h]
    \centering
    \includegraphics[trim={6cm 22.7cm 5cm 3cm},clip,width=0.4\textwidth]{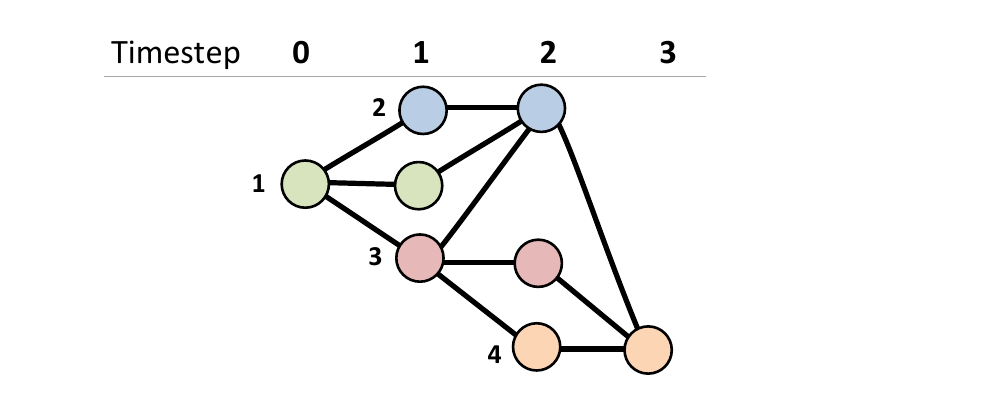}
    \vspace{-0.2cm}\caption{An example of MDD expansion for token of color $c_1$.}
    \label{figure-MDD}
\end{figure}

An example of MDD expansion for orange token ($c_1$) from Figure \ref{figure-TSWAP} in shown in Figure \ref{figure-MDD}.

\section{Experimental Evaluation}
We have implemented suggested modifications of CBS algorithm and MDD-SAT solver. The CBS algorithm for TSWAP has been implemented from scratch in C++. To obtain MDD-SAT applicable on TSWAP we modified the existing C++ implementation. All experiments were run on an i7 CPU 2.6 Ghz under Kubuntu linux 16 with 8GB RAM. \footnote{To enable reproducibility of presented results we provide complete source code of our TSWAP solvers on: \texttt{http://users.fit.cvut.cz/ $\sim$surynpav/research/ictai2018}.}

\begin{figure}[h]
    \centering
    \includegraphics[trim={4cm 25cm 11cm 3.3cm},clip,width=0.4\textwidth]{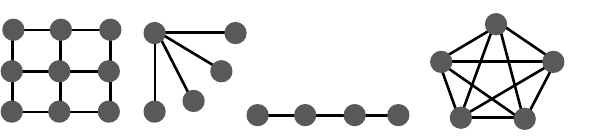}
    \vspace{-0.2cm}\caption{Example of regular 4-connected {\em grid}, {\em star}, {\em path/bubblesort}, and {\em clique}.}
    \label{figure-types}
\end{figure}

The experimental evaluation has been done on diverse instances consisting of grid graphs of sizes $8 \times 8$ and $16 \times 16$, random graphs containing 50\% of random edges, star graphs, path graphs and cliques (see \ref{figure-types}). Initial and goal configurations of agents and tokens have been generated randomly in all tests. The tested TSWAP instances had always all tokens of different colors and the graph was always fully occupied by tokens. On the other hand in MAPF tests, we varied occupancy of the graph.

A TSWAP instance corresponding to a not fully occupied MAPF instance has been obtained by interpreting empty positions as a separate color. It is important to note that presented algorithms are applicable for TSWAP instances with multiple tokens having the same color.

\subsection{Comparison TSWAP Algorithms}

Our first test has been focused on comparison of modified CBS and MDD-SAT on TSWAP instances. In this case we used random graphs, stars, paths, and cliques consisting of 4 to 16 vertices. For each graph size we generated 10 instances and measured runtime of CBS and MDD-SAT. he timeout in all tests has been set to 300 seconds. The number of swaps has been collected too.

\begin{figure}[h]
    \centering
    \includegraphics[trim={2.5cm 20.5cm 2cm 2.5cm},clip,width=0.5\textwidth]{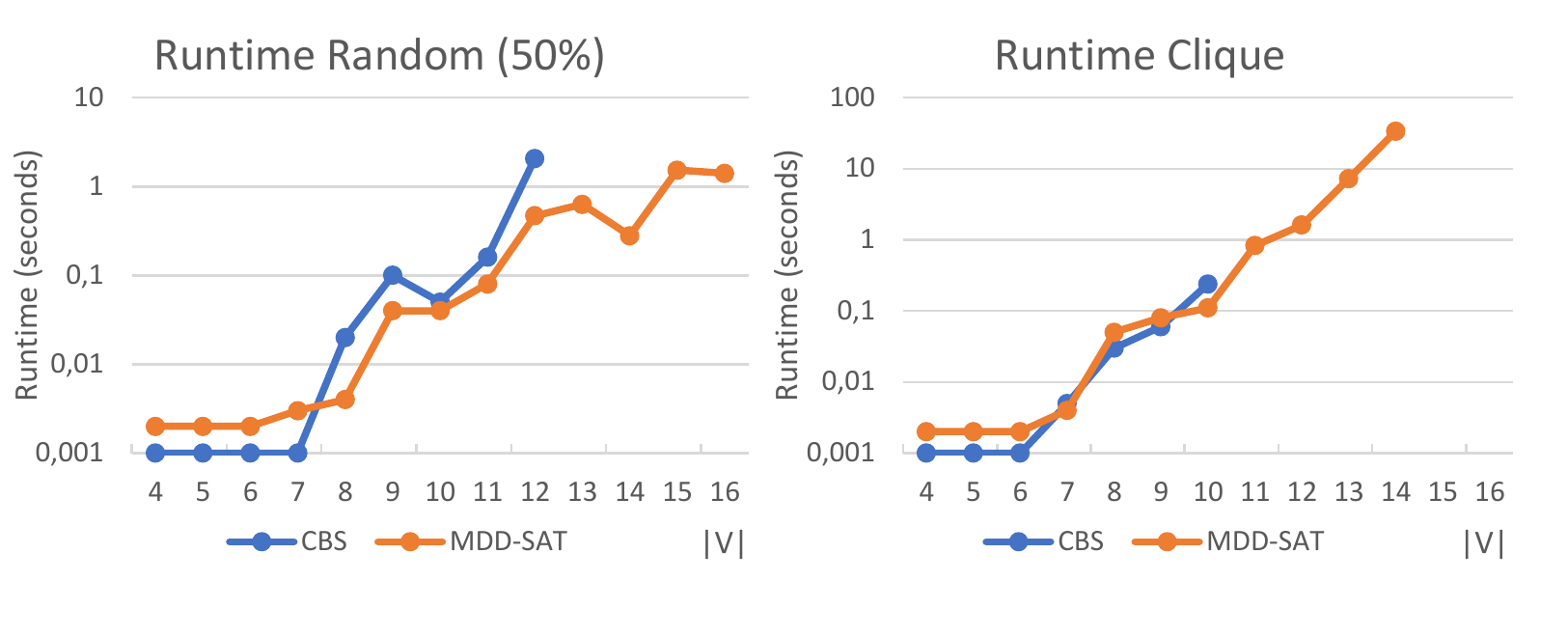}
    \vspace{-1.0cm}\caption{Comparison of modified CBS and MDD-SAT algorithms on TSWAP instances on random graphs containing 50\% of random edges and on cliques.}
    \label{exp-random-clique}
\end{figure}

\begin{table}[h]
\centering
\def\arraystretch{1.1}
\caption{Mean number of swaps on random graphs (50\%) and cliques}
\label{tbl-random-clique}
\begin{tabularx}{0.40\textwidth}{l|lllllllllllll}
$|V|$              & 4  &  6 & 8 & 10 & 12 & 14 & 16 \\ \hline
Random(50\%) & 9 & 11 & 14 & 27 & 26 & 33 & 39 \\ \hline
Clique              & 9 & 11 & 14 & 27 & 30 & 38 & 49
\end{tabularx}
\end{table}

\begin{figure}[h]
    \centering
    \includegraphics[trim={2.5cm 20.5cm 2cm 2.5cm},clip,width=0.5\textwidth]{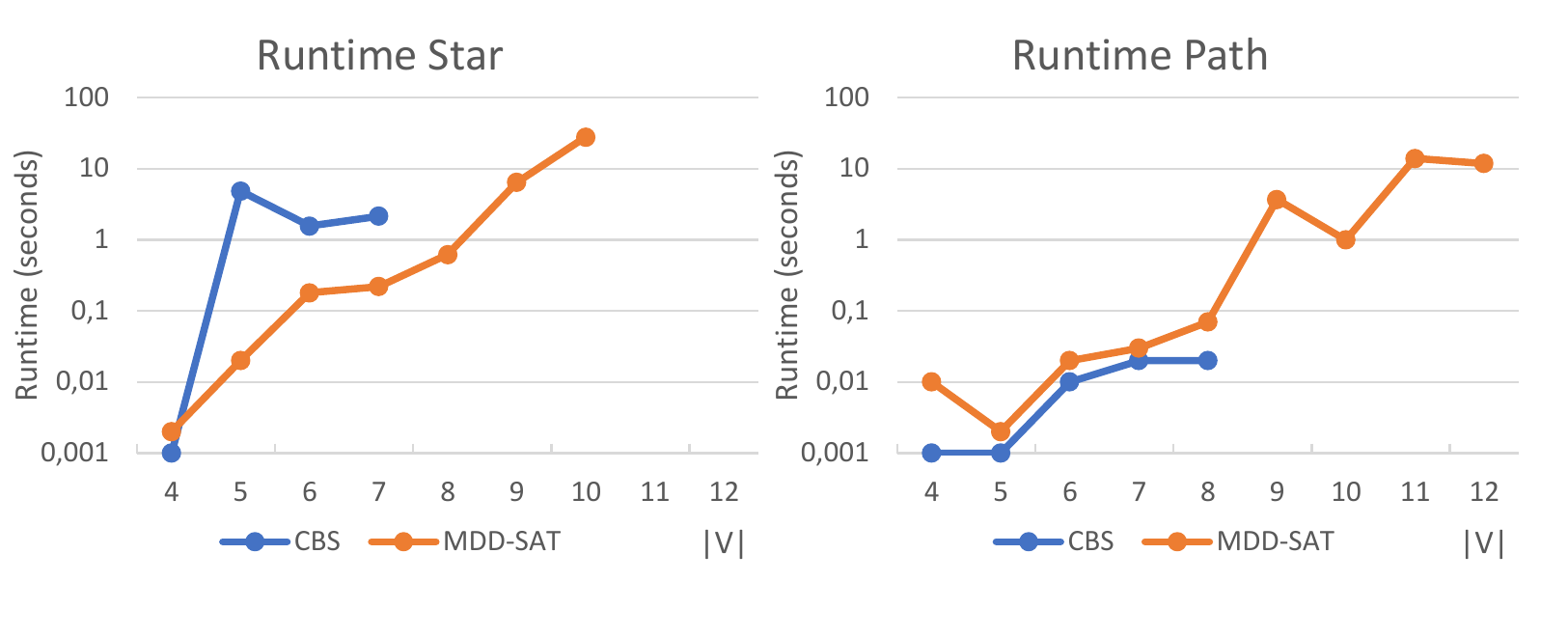}
    \vspace{-1.0cm}\caption{Comparison of modified CBS and MDD-SAT algorithms on TSWAP instances on path and star graphs.}
    \label{exp-star-path}
\end{figure}

Mean values for runtime are presented in Figures \ref{exp-random-clique} and \ref{exp-star-path}. For smaller graphs the performance comparison of CBS and MDD-SAT has no clear winner however for larger graphs MDD-SAT tends to dominate. We attribute this to worse performance of CBS when dealing with too many conflicts which is a typical property of TSWAP problem. The mean number of swaps are shown in Tables \ref{tbl-random-clique} and \ref{tbl-star-path}.

\begin{table}[h]
\centering
\def\arraystretch{1.1}
\caption{Mean number of swaps on star and paths graphs}
\label{tbl-star-path}
\begin{tabularx}{0.40\textwidth}{l|lllllllllllll}
$|V|$  & 4  &  5 & 6 & 7 & 8 & 9 & 10 & 11 & 12 \\ \hline
Star    & 9 & 34 & 32 & 34 & 38 & 54 & 66 & - & - \\ \hline
Path   & 20 & 14 & 26 & 35 & 33 & 64 & 47 & 94 & 97
\end{tabularx}
\end{table}

\subsection{Comparison of Difficulty of TSWAP and MAPF}

\begin{figure}[h]
    \centering
    \includegraphics[trim={2cm 21cm 4cm 2.5cm},clip,width=0.5\textwidth]{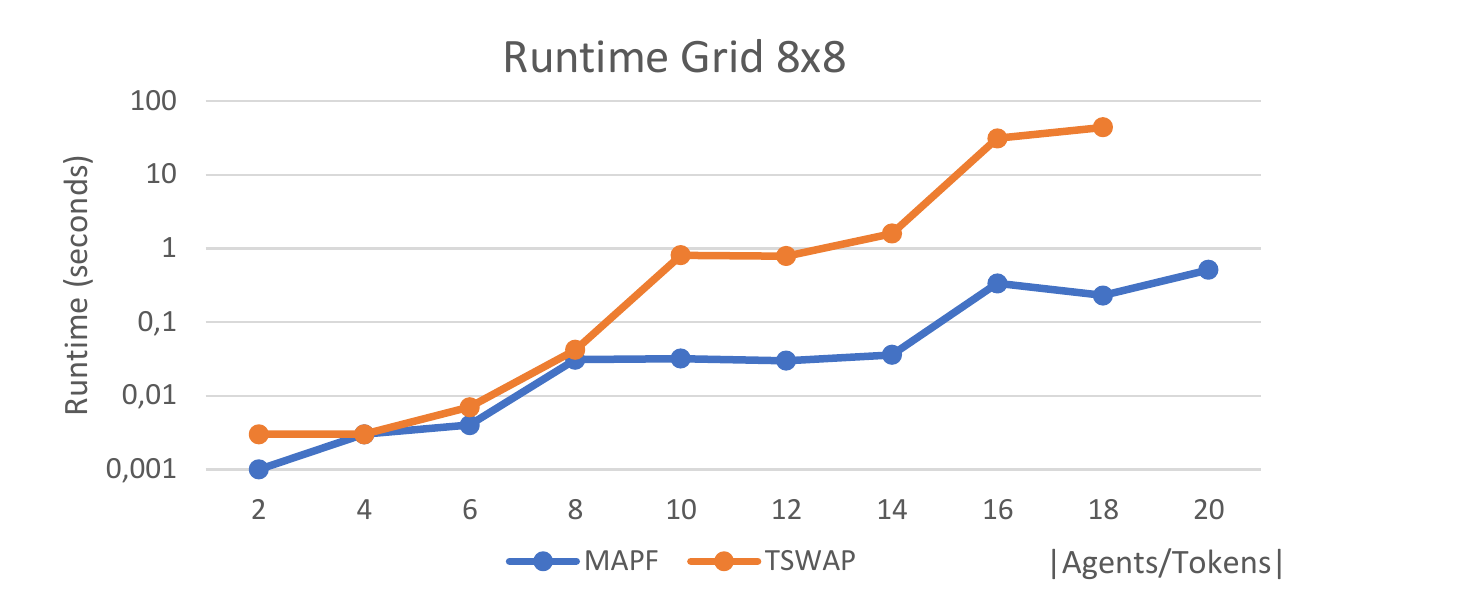}
    \vspace{-0.5cm}\caption{Comparison of MAPF and TSWAP solving on a grid of size $8 \times 8$.}
    \label{exp-grid8x8}
\end{figure}

\begin{figure}[t]
    \centering
    \includegraphics[trim={2cm 21cm 4cm 2.5cm},clip,width=0.5\textwidth]{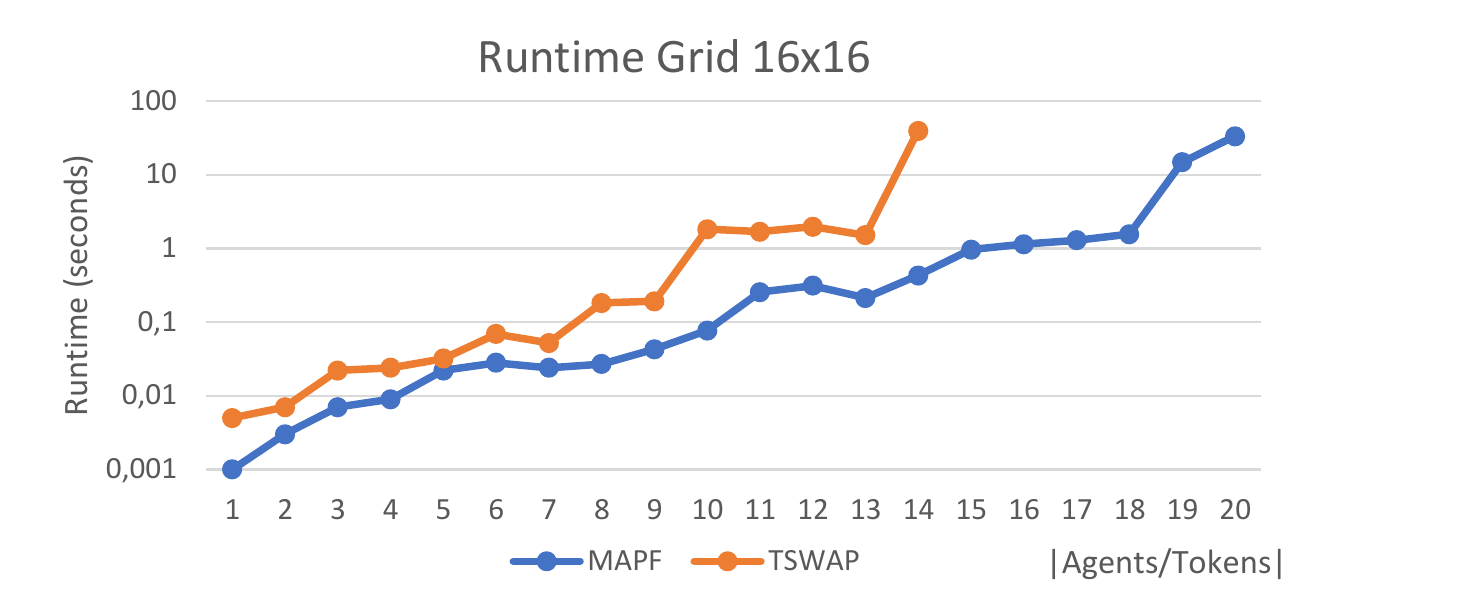}
    \vspace{-0.5cm}\caption{Comparison of MAPF and TSWAP solving on a grid of size $16 \times 16$.}
    \label{exp-grid16x16}
\end{figure}

We also compared difficulty of TSWAP and MAPF. This has been done only with MDD-SAT as it has proved better performance in the previous test. We used grids $8 \times 8$ and $16 \times 16$ in this test. For a fixed graph we gradually increased the number of tokens/agents and measured the runtime. For each number of agents/tokens we generated 10 random instances. Mean values are reported in Figures \ref{exp-grid8x8} and \ref{exp-grid16x16}.

The evaluation on grids shows almost the order of magnitute speedup for the MAPF problem compared to TSWAP. This is a surprising result as theoretically TSWAP requires fewer swaps than MAPF moves.

\section{Conclusion}

We have observed similarities between two problems being addressed in computer science - the {\em multi-agent path finding} (MAPF) and {\em token swapping problem} (TSWAP). We focused on optimal TSWAP solving. We demonstrated how to modify existing MAPF algorithms CBS and MDD-SAT for TSWAP. To our best knowledge this is the first attempt to solve TSWAP optimally from the practical solving point of view.

Although we performed our tests on a limited set of benchmarks for this short paper, our experimental results indicate that modifications of CBS and MDD-SAT represent a viable options for TSWAP solving while MDD-SAT exhibits better performance. Results also indicate that MAPF problem is easier than TSWAP.

For the future work we plan to evaluate TSWAP solving on a more diverse set of benchmarks. As TSWAP permits multiple tokens to have the same color, we plan to investigate usage of meta-tokens consisting of tokens of the same color whose shortest paths will be found as network flows \cite{DBLP:conf/atal/MaK16}.


\bibliographystyle{IEEEtran}
\bibliography{bibfile}

\vfill
\end{document}